\documentclass[11pt]{article}

\usepackage[utf8]{inputenc}
\usepackage[T1]{fontenc}
\usepackage{amsmath}
\usepackage{amsfonts}
\usepackage{amssymb}
\usepackage{graphicx}
\usepackage{booktabs}
\usepackage{tabularx}
\usepackage{multirow}
\usepackage{siunitx}
\usepackage{url}
\usepackage{hyperref}

\usepackage[margin=1in]{geometry}
\usepackage{setspace}
\usepackage[numbers,sort&compress]{natbib}

\hypersetup{
    colorlinks=true,
    linkcolor=blue,
    filecolor=magenta,      
    urlcolor=cyan,
    citecolor=blue,
}

\begin{document}

\title{Controllable Probabilistic Forecasting with Stochastic Decomposition Layers}

\author{
John S. Schreck$^{1,*}$, 
William E. Chapman$^{2}$, 
Charlie Becker$^{1}$, \\
David John Gagne II$^{1}$, 
Dhamma Kimpara$^{1}$, 
Nihanth Cherukuru$^{1}$, \\
Judith Berner$^{1}$, 
Kirsten J. Mayer$^{1}$, 
Negin Sobhani$^{1}$ \\[0.3cm]
\small $^{1}$NSF National Center for Atmospheric Research, Boulder, CO, USA \\
\small $^{2}$Atmospheric and Oceanic Sciences Department, University of Colorado, Boulder, CO, USA \\
\small $^{*}$Corresponding author: schreck@ucar.edu
}

\date{\today}

\maketitle

\begin{abstract}
AI weather prediction ensembles with latent noise injection and optimized with the continuous ranked probability score (CRPS) have produced both accurate and well-calibrated predictions with far less computational cost compared with diffusion-based methods. However, current CRPS ensemble approaches vary in their training strategies and noise injection mechanisms, with most injecting noise globally throughout the network via conditional normalization. This structure increases training expense and limits the physical interpretability of the stochastic perturbations. We introduce Stochastic Decomposition Layers (SDL) for converting deterministic machine learning weather models into probabilistic ensemble systems. Adapted from StyleGAN's hierarchical noise injection, SDL applies learned perturbations at three decoder scales through latent-driven modulation, per-pixel noise, and channel scaling. When applied to WXFormer via transfer learning, SDL requires less than 2\% of the computational cost needed to train the baseline model. Each ensemble member is generated from a compact latent tensor (5 MB), enabling perfect reproducibility and post-inference spread adjustment through latent rescaling. Evaluation on 2022 ERA5 reanalysis shows ensembles with spread-skill ratios approaching unity and rank histograms that progressively flatten toward uniformity through medium-range forecasts, achieving calibration competitive with operational IFS-ENS. Multi-scale experiments reveal hierarchical uncertainty: coarse layers modulate synoptic patterns while fine layers control mesoscale variability. The explicit latent parameterization provides interpretable uncertainty quantification for operational forecasting and climate applications.
\end{abstract}

\section{Introduction}
National Meteorological and Hydrological Services have increasingly incorporated ensemble forecasting methods to provide probabilistic predictions and quantify forecast uncertainty over the past 3 decades \cite{leutbecher2008svs, palmer2019stochastic}. Traditional ensemble numerical weather prediction (NWP) approaches perturb initial conditions and/or use perturbed model physics parameterizations \cite{molteni1996ens,buizza2005comparison,leutbecher2017stochastic, berner2017stochastic}, but these methods are computationally expensive, requiring multiple full model integrations. Operational NWP ensembles have a relatively small number of members, usually on the order of 30-50, and may run at a lower spatial resolution compared with deterministic flagship members \cite{Fu2024-zo}. Recent advances in machine learning for weather prediction \cite{keisler2022fourcastnet, pathak2022fourcastnet, bi2023pangu, Lam2023graphcast, chen2023fuxi, schreck2025} have demonstrated skill comparable to traditional numerical weather prediction models, yet extending these capabilities to probabilistic forecasting while respecting chaotic dynamics remains challenging. 

The theoretical foundation for ensemble prediction stems from Lorenz's discovery that the atmosphere is a chaotic system where small initial condition errors grow exponentially \cite{lorenz1963deterministic}. Quantitative estimates from early global circulation models established a practical deterministic predictability limit of roughly two weeks \cite{charney1966applicability}, later confirmed as a fundamental property of atmospheric dynamics \cite{lorenz1969predictability, Lorenz1969-tc, lorenz1996predictability}. Epstein and coauthors \cite{Epstein1969-nj} proposed a stochastic-dynamic formulation of the primitive equations that predicted the mean and variance of each state variable but was viewed to be computationally intractable at the time. Monte Carlo ensemble approximations \cite{Leith1974} and optimization with least squares techniques \cite{Pitcher1977-li} demonstrated that calibrated NWP ensembles could be produced that could improve on deterministic forecasts. In general, ensemble forecasting quantifies the fundamental predictability limitations of chaotic dynamics by sampling the phase space of possible atmospheric states, representing forecast uncertainty through trajectory divergence on the climate system's attractor \cite{palmer2019stochastic, toth1993ensemble}.

Early ML weather models employed deterministic architectures trained with mean squared error (MSE) loss. When combined with multi-step training, these models produced overly smooth forecasts that underrepresented atmospheric variability. Generative modeling approaches, particularly diffusion-based weather models \cite{sohl-dickstein2015deep, ho2020denoising, song2021scorebased, price2025gencast}, address this limitation by iteratively denoising random fields to produce realistic ensemble members. However, diffusion-based weather models \cite{price2025gencast} require learning full denoising trajectories across hundreds of timesteps during training and at least 20-50 iterative refinement steps per forecast at inference, imposing substantial computational burden on operational systems requiring rapid generation of large ensembles (50+ members).

Recent work has explored parameter-efficient alternatives that avoid iterative denoising by injecting stochasticity through various architectural mechanisms and training directly on the Continuous Ranked Probability Score (CRPS) to achieve calibrated ensembles with computational costs proportional to ensemble size \cite{lang2024aifs, alet2024fgn, bonev2025fourcastnet3, fuxi_ens, kochkov2024neuralgcm}. Foundation models demonstrate strong transfer learning capabilities across weather and climate tasks \cite{nguyen2023climax, bodnar2025aurora, allen2025aardvark}. Alternative approaches to ensemble generation include conditional diffusion models with explicit uncertainty quantification \cite{shi2025codicast}, latent diffusion with knowledge alignment mechanisms for precipitation nowcasting \cite{gao2023prediff}, and diffusion-based ensemble emulation methods that generate large ensembles conditioned on few operational forecast members \cite{price2024seeds}. From a dynamical systems perspective, these approaches sample the forecast distribution but provide no mechanism to systematically navigate the sampled trajectories or regenerate specific ensemble members. In addition, many of these approaches inject noise globally throughout the network without explicit hierarchical decomposition, which limits the spatial and scale-specific interpretability of the stochastic perturbations.  

Operational ensemble forecasting requires not only probabilistic skill but also interpretable control over uncertainty structure. Forecasters benefit from the ability to systematically explore alternative scenarios within the ensemble distribution. For example, amplifying or suppressing specific synoptic patterns to understand their downstream impacts on regional weather. Similarly, ensemble post-processing and recalibration workflows require adjusting spread without regenerating forecasts, particularly when computational constraints limit ensemble size or when verification reveals systematic over- or underdispersion \cite{hamill1997verification, vannitsem2021statistical}. The multi-scale nature of atmospheric predictability, where large-scale synoptic systems exhibit different error growth rates than mesoscale features \cite{tribbia2004scale}, motivates hierarchical uncertainty decomposition. Injecting perturbations at distinct spatial scales spanning the synoptic/meso-$\alpha$ boundary through the meso-$\beta$ range while maintaining meteorological consistency enables physically interpretable ensemble spread that reflects the scale-dependent structure of forecast uncertainty. Current ML ensemble methods lack explicit architectural mechanisms for scale-dependent uncertainty control and provide limited capability to adjust or explore the generated forecast distribution after inference.

In this work, we introduce Stochastic Decomposition Layers (SDL) that separate latent-driven modulation from per-pixel stochasticity, enabling controlled exploration of forecast uncertainty. We build upon the WXFormer architecture \cite{schreck2025, chapman2025camulator}, incorporating SDL at multiple hierarchical scales to enable two critical capabilities: (1) calibrated probabilistic forecasts competitive with physics-based ensembles, achieved through CRPS-based training with computational efficiency comparable to other computationally efficient methods, and (2) post-inference exploration of forecast uncertainty by systematically perturbing latent tensor and noise components. While noise scaling is theoretically possible in any stochastic model, SDL learns a symmetric latent geometry that provides calibrated control over ensemble spread without regeneration. Each ensemble member is determined by a compact latent tensor, enabling perfect reproducibility and spread adjustment after inference. This provides operational control over forecast uncertainty without regenerating ensemble members.

\section{Methods}

\subsection{Base Architecture}

\begin{figure}[htbp]
\centering
\includegraphics[width=\textwidth]{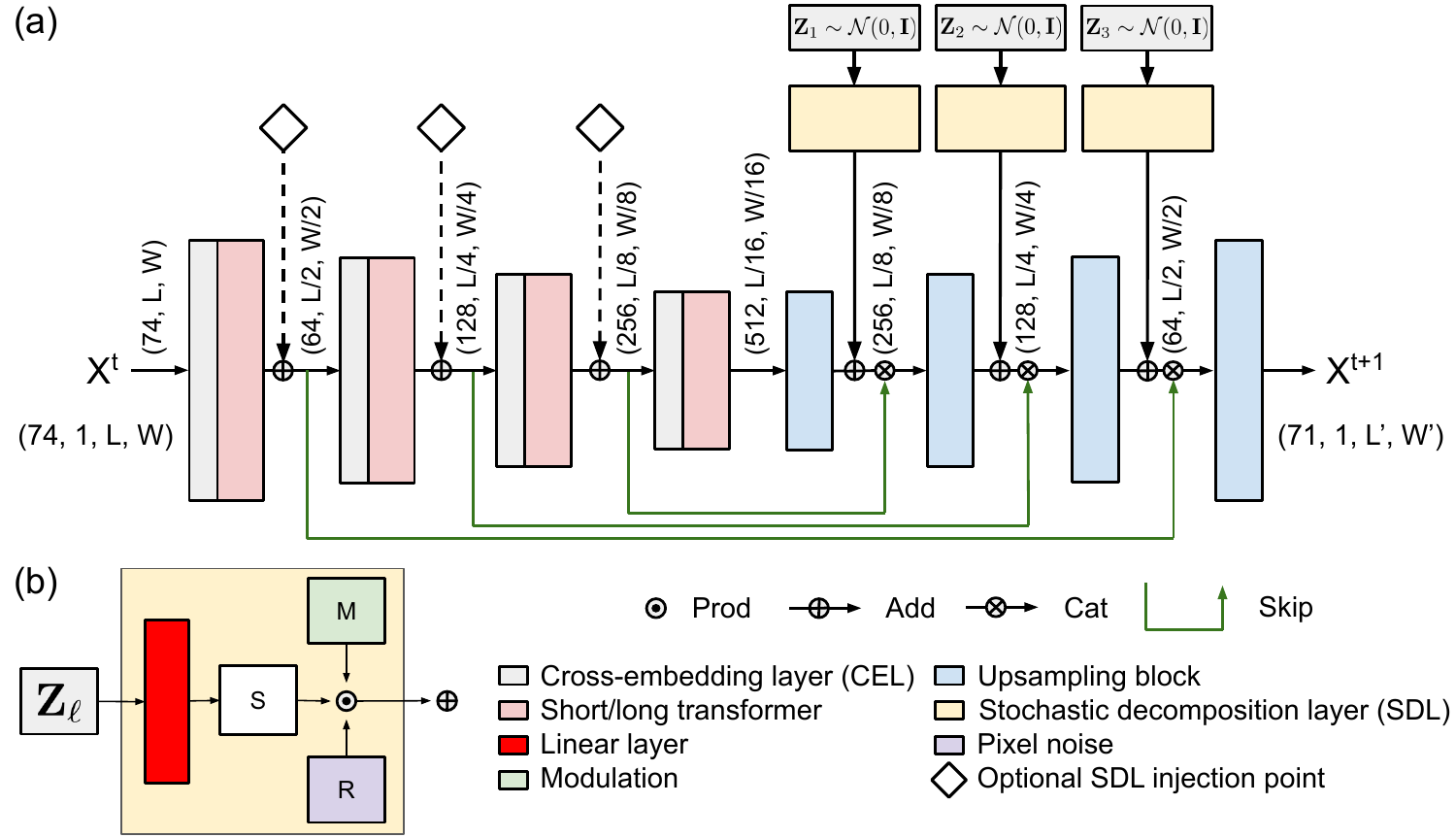}
\caption{Schematic of WXFormer with SDL injection points. Independent latent vectors $Z_1$, $Z_2$, and $Z_3$ are sampled from $\mathcal{N}(0,\mathbf{I})$ and fed to each of the three decoder SDL layers (yellow boxes) after upsampling blocks and before skip connection concatenation. Diamonds mark optional encoder SDL injection points not used in this work. Each SDL operates with independent learnable parameters to accommodate feature magnitude variations across network depth. (b) Data flow through the SDL. Latent vector $Z_\ell$ is transformed via a learned linear layer (red) to produce  tensor $S$. Three independent pathways (learned channel modulation at the top [$\mathbf{M}$], latent ``style'' tensor in the middle [$\mathbf{S}$], and per pixel noise generation at the bottom [$\mathbf{R}$]) are combined multiplicatively before the residual addition to the input features. Tensor dimensions are shown for clarity; spatial broadcasting expands $\mathbf{S}$ and $\mathbf{M}$ to match $\mathbf{R}$}
\label{fig:wxformer_noise}
\end{figure}

The WXFormer model \cite{schreck2025, chapman2025camulator, sha2025conservation} employs a hierarchical encoder-decoder structure for weather prediction (Figure~\ref{fig:wxformer_noise}). The encoder processes input features on progressively coarser spatial grids through a series of cross embedding layers (CEL) and transformer blocks, enabling efficient multi-scale feature extraction. The decoder uses U-Net-style skip connections to integrate high-resolution spatial information from the encoder with globally-informed features from the deepest layer, producing weather forecasts at the original input resolution.

This multi-scale processing provides injection points for stochasticity across different atmospheric resolutions, from coarse large-scale structures in deeper layers to fine-grained spatial details in shallower layers. By strategically placing noise injection layers at each level of this hierarchy (Figure~\ref{fig:wxformer_noise}), we enable simultaneous perturbation of large-scale circulation patterns and small-scale atmospheric features, creating ensemble members that span a physically coherent range of forecast solutions.

\subsection{Stochastic Decomposition Layer}
Recent approaches to ensemble generation in machine learning weather models typically apply noise globally without explicit scale separation. The StyleGAN architecture \cite{karras2019style, karras2020analyzing} demonstrated that hierarchical, decomposed noise injection enables both diversity and fine grained control in generative models by separating global style variations, which influence composition and large scale structure, from local stochastic details, which influence texture and fine scale features. Style transfer techniques have been successfully adapted to atmospheric science applications \cite{schreck2023style}, showing that neural style decomposition can capture complex physical relationships in atmospheric data.

\begin{table}[t!]
\centering
\begin{tabular}{l l l}
\toprule
\textbf{Component} & \textbf{Operation} & \textbf{Shape} \\
\midrule
\texttt{feature\_map} & input features & $(B, C, H, W)$ \\
\texttt{noise} & latent input & $(B, \text{noise\_dim})$ \\
$\mathbf{S}$ & latent-derived ``style'' & $(B, C, 1, 1)$ \\
$\mathbf{M}$ & learned channel modulation & $(1, C, 1, 1)$ \\
$\mathbf{R}$ & per-pixel Gaussian noise & $(B, C, H, W)$ \\
$\alpha$ & fixed noise scaling & scalar \\
output & $\mathbf{F}_{\text{in}} + \alpha \cdot \mathbf{R} \odot \mathbf{S} \odot \mathbf{M}$ & $(B, C, H, W)$ \\
\bottomrule
\end{tabular}
\caption{Components of the SDL layer corresponding to Equation~\ref{eq:sdl} and Figure~\ref{fig:wxformer_noise}b.}
\label{tab:tensor_shapes}
\end{table}

These insights motivate our adaptation of hierarchical noise decomposition for atmospheric ensemble forecasting. We accomplish this by injecting controlled stochastic perturbations into feature maps using three multiplicative components within SDL. The output feature map $\mathbf{F}_{\text{out}} \in \mathbb{R}^{B \times C \times H \times W}$ is computed as:
\begin{equation}
\mathbf{F}_{\text{out}} = \mathbf{F}_{\text{in}} + \alpha \cdot \mathbf{R}_{\text{Gauss}}  \odot \mathbf{S} \odot \mathbf{M}
\label{eq:sdl}
\end{equation}
where $\mathbf{R}_{\text{Gauss}} \sim \mathcal{N}(0, \mathbf{I})$ is per-pixel Gaussian noise sampled at full spatial resolution, $\alpha$ is a non-trainable scaling parameter, $\mathbf{S} \in \mathbb{R}^{B \times C \times 1 \times 1}$ is a tensor derived from a latent tensor $\mathbf{Z}_\ell \sim \mathcal{N}(0, \mathbf{I})$ sampled independently at each SDL layer via learned linear transformation, and $\mathbf{M} \in \mathbb{R}^{1 \times C \times 1 \times 1}$ is a learnable channel-wise modulation parameter. The operator $\odot$ denotes element-wise multiplication with broadcasting across spatial dimensions. Table~\ref{tab:tensor_shapes} details the tensor operations and shape transformations implementing this layer. Unlike StyleGAN's original implementation which uses correlated noise across hierarchical levels, we sample independent latent vectors $\mathbf{Z}_\ell \sim \mathcal{N}(0, \mathbf{I})$ at each SDL injection point to allow distinct uncertainty representations at different scales, where synoptic/meso-$\alpha$ (400-800~km) and meso-$\beta$ (100-200~km) processes have different predictability characteristics. The multiplicative structure in Equation~\ref{eq:sdl} enables independent control over ensemble characteristics: the linear transformation producing $\mathbf{S}$ and the channel modulation parameter $\mathbf{M}$ are learned during training, while $\alpha$ remains fixed to provide consistent noise scaling. During inference, sampling different $\mathbf{Z}$ tensors produces distinct ensemble members, with the learned parameters ensuring perturbations remain appropriately scaled and physically plausible.



\subsection{Multi-Scale Injection Strategy}
The SDL architecture supports injection at six hierarchical levels: three encoder positions and three decoder positions (Figure~\ref{fig:wxformer_noise}). In this work, we implement decoder-only SDL injection with three independent latent vectors $\mathbf{Z}_1$, $\mathbf{Z}_2$, $\mathbf{Z}_3$ sampled at each decoder stage (Figure~\ref{fig:wxformer_noise}, yellow boxes). In the decoder, SDL layers are positioned after each upsampling operation but before skip connection concatenation, ensuring that stochastic perturbations are applied to upsampled features before incorporating encoder information. This ordering enables localized variations while maintaining large-scale consistency through the skip connections.

\begin{table}[h!]
\centering
\small
\begin{tabular}{lccccp{5.0cm}}
\toprule
\textbf{Level} & \textbf{Grid} & \textbf{Resolution (°)} & \textbf{Scale (zonal km)} & \textbf{SDL} & \textbf{Scale interpretation} \\
 & (lat×lon) & ($\Delta\phi$ × $\Delta\lambda$) & (@45°N) & \textbf{layer} & \\
\midrule
1 & 24×36   & 7.5×10.0  & 787 &
$\beta_1$ &
Synoptic/meso-$\alpha$ boundary scale. Modulation of continental-scale circulation patterns, broad troughs and ridges, and large-scale baroclinic wave structure. \\
2 & 48×72   & 3.75×5.0  & 394 &
$\beta_2$ &
Meso-$\alpha$ scale. Regional cyclone structure, frontal systems, and mesoscale organization of synoptic disturbances. \\
3 & 96×144  & 1.875×2.5 & 197 &
$\beta_3$ &
Meso-$\alpha$/meso-$\beta$ transition scale. Mesoscale convective organization, frontal bands, and regional precipitation structure. \\
4 & 192×288 & 0.94×1.25 & 98  &
output &
Meso-$\beta$ scale. Mesoscale convective systems and frontal gradients. \\
\bottomrule
\end{tabular}
\caption{Effective spatial resolution and atmospheric scale interpretation across U-Net decoder levels. Scale shown for longitude at 45°N (representative mid-latitude). SDL perturbations are injected at Levels~1--3 (bottleneck through second upsampling) via parameters $\beta_1$, $\beta_2$, $\beta_3$ respectively. Level~4 is the final output layer (no SDL injection).}
\label{tab:decoder_scales}
\end{table}


To enable extensive experimentation within computational constraints, we implement decoder-only SDL injection with independent latent sampling at each layer. We refer to architectural positions as ``decoder levels'' (Level~1 through Level~4, Table~\ref{tab:decoder_scales}) and SDL application locations as ``injection points'' controlled by parameters $\beta_1$, $\beta_2$, and $\beta_3$ respectively, spanning from the synoptic/meso-$\alpha$ boundary (Level~1) through the meso-$\beta$ scale (Level~4).

The decoder-only configuration requires storing three latent tensors per ensemble member (one per injection point), totaling approximately 5 MB in single precision (float32): Level 1 (24$\times$36$\times$71, $\sim$240 KB), Level 2 (48$\times$72$\times$71, $\sim$960 KB), and Level 3 (96$\times$144$\times$71, $\sim$3.7 MB). Full encoder-decoder injection would require six tensors ($\sim$10 MB per member). These compact latent representations enable perfect reproducibility of any ensemble member and support post-inference spread adjustment through latent rescaling, all while requiring orders of magnitude less storage than archiving complete forecast fields.

Each injection point employs a fixed scaling parameter $\alpha$ to accommodate feature magnitude variations across network depth. Appropriate $\alpha$ selection enables reasonable initial ensemble spread, accelerating convergence of learnable SDL components ($\mathbf{M}$ and the linear transformation generating $\mathbf{S}$). The transfer learning framework allows freezing pre-trained WXFormer weights while training only SDL parameters, or joint fine-tuning for superior ensemble quality. We set $\alpha = 0.235$ at all injection points based on preliminary experiments.

\subsection{Ensemble Generation}
During inference, multiple ensemble members are generated by sampling different latent vectors $\{\mathbf{Z}_1, \mathbf{Z}_2, \mathbf{Z}_3\}$ independently at each of the three decoder SDL injection points. The learned parameters $\mathbf{M}$ and the linear transformations producing $\mathbf{S}$, combined with the fixed $\alpha$ scaling parameters, ensure variations remain physically plausible and appropriately scaled. This multi-scale injection produces comprehensive uncertainty quantification: deep-layer perturbations affect large-scale flow patterns (e.g., synoptic troughs and ridges), while shallow-layer perturbations modulate fine-scale features (e.g., mesoscale convective systems and surface temperature gradients). The ensemble spread reflects the model's learned uncertainty structure rather than relying on ad-hoc perturbation strategies.

\section{Training dataset}
This study uses ERA5 reanalysis \cite{hersbach2020era5} coarsened to 1$^\circ$ horizontal resolution (360$\times$181 grid) spanning 1979-2022. We selected 16 vertical levels from ERA5's 137 model levels (levels 10, 30, 40, 50, 60, 70, 80, 90, 95, 100, 105, 110, 120, 130, 136, 137), along with surface variables, static fields (land-sea mask, surface geopotential), and instantaneous solar irradiance as forcing, yielding 74 total input variables. This vertical discretization emphasizes lower-to-mid tropospheric representation where most weather-relevant processes occur while accepting coarser resolution at higher altitudes. The temporal split follows \cite{schreck2025}: training (1979-2017), validation (2018-2019), and testing (2020-2022). Complete details on variable selection, vertical level specification, and data processing are provided in \cite{schreck2025, sha2025terrain}.

\section{Fine-Tuned Training Scheme}


We initialize from a pre-trained deterministic WXFormer model trained on ERA5 following \cite{schreck2025}. The SDL components ($\alpha$, $\mathbf{M}$, and the linear transformations producing $\mathbf{S}$) are inserted into the model, and all model weights are then fine-tuned jointly. This transfer learning approach leverages the pre-trained model's deterministic forecast skill as a strong initialization, requiring approximately 1.61\% of the computational cost needed to train the baseline WXFormer from scratch (see ref \cite{schreck2025} for baseline training details). Fine-tuning all weights produces superior ensemble quality compared to training only the SDL parameters while maintaining computational efficiency. See ~\ref{sec:train_details} for complete training details and computational cost analysis.

During training, we generate $M$ = 10 ensemble members per input by sampling different latent tensors $\mathbf{Z}$ at each SDL injection point. The SDL parameters are optimized using the almost fair Continuous Ranked Probability Score (afCRPS) loss \cite{lang2024aifs}, defined for an ensemble $\{x_j\}_{j=1}^M$ and verification target $y$ as:

\begin{equation}
\text{afCRPS}_\alpha = \frac{1}{M}\sum_{j=1}^M |x_j - y| - \frac{1-\epsilon}{2M(M-1)}\sum_{j=1}^M\sum_{k=1}^M |x_j - x_k|
\label{eq:afcrps}
\end{equation}

where $\epsilon := \frac{(1-\alpha)}{M}$, and $\alpha \in (0,1]$ is a hyperparameter. Setting $\alpha = 1$ recovers the fair CRPS \cite{ferro2013fair}, which adjusts for finite ensemble size. However, fair CRPS suffers from a degeneracy where all members except one match the target, leaving the remaining member unconstrained. The almost fair formulation with $\alpha < 1$ avoids this pathology while maintaining near-fair scoring properties. We use $\alpha = 0.95$ following \cite{lang2024aifs}. The loss is averaged across all variables, vertical levels, and spatial locations with latitude weighting following \cite{schreck2025}.

Training uses the same 1979-2017 ERA5 data as the deterministic model, with validation performed on 2018-2019 and testing on 00Z initializations from 2022. This fine-tuning strategy is substantially more efficient than diffusion-based ensemble generation, which requires iterative denoising steps at inference and prohibitively expensive training on full trajectory sequences. Our approach incurs computational overhead only proportional to the desired ensemble size. Generating 50 members requires 50 forward passes with different $\mathbf{Z}$ samples but no architectural modifications or iterative refinement. The pretrained base model ensures that deterministic forecast quality is preserved without training from scratch, while the learned SDL parameters add controlled stochasticity appropriate to each hierarchical level.

\section{Results}

\subsection{Ensemble Forecast Skill}
\label{sec:skill}

\begin{figure}[htbp]
    \centering
    \includegraphics[width=\columnwidth]{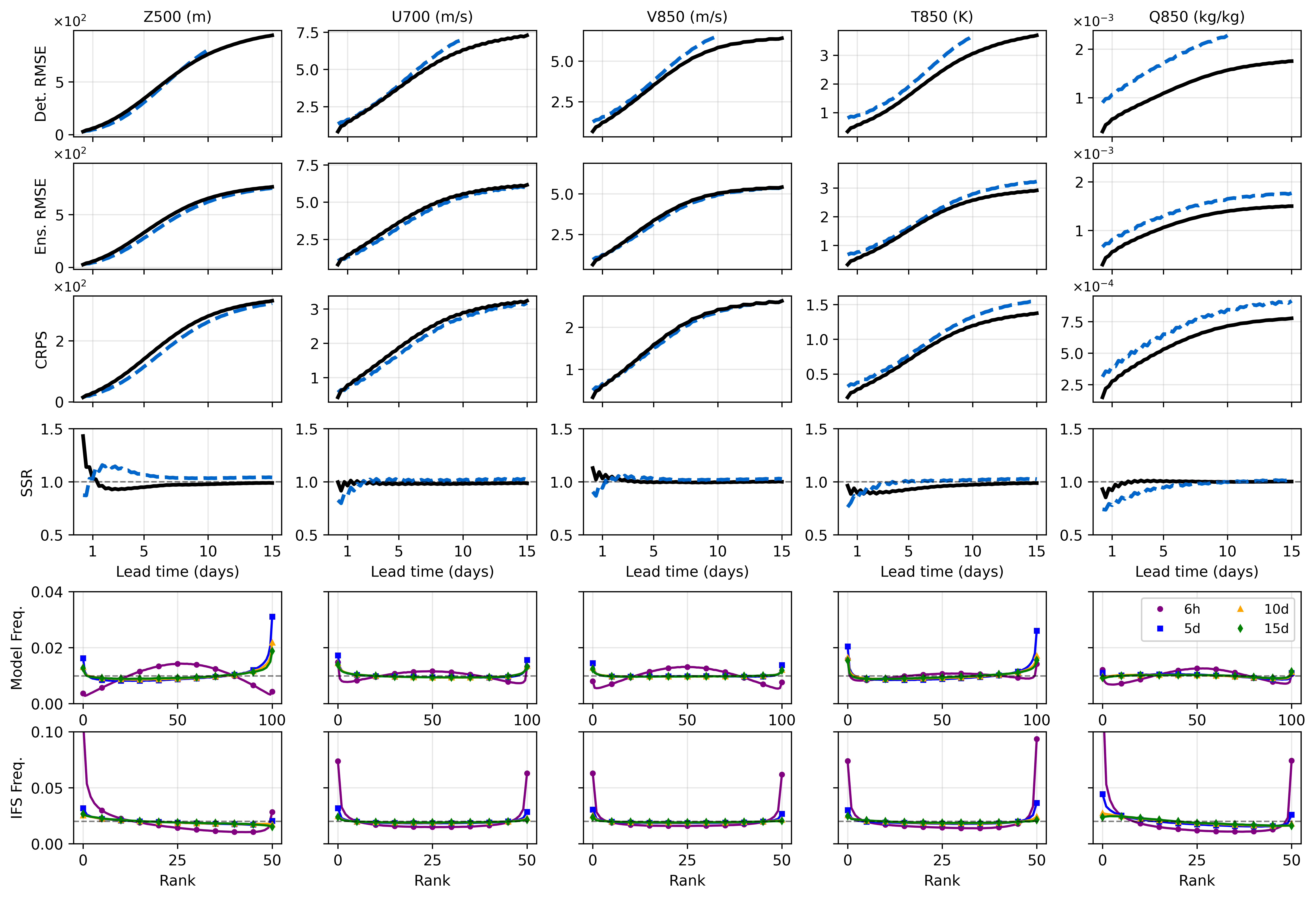}
    \caption{Forecast verification metrics for selected atmospheric variables at key pressure levels over the 2022 ERA5 test period. Columns show geopotential height at 500~hPa (Z500), zonal wind at 700~hPa (U700), meridional wind at 850~hPa (V850), temperature at 850~hPa (T850), and specific humidity at 850~hPa (Q850). Rows display: (1) Deterministic Root Mean Square Error (RMSE) comparing the baseline deterministic WXFormer (solid line, 15-day forecasts) against IFS HRES (dashed line, 10-day forecasts); (2) Ensemble mean RMSE comparing SDL-WXFormer (solid line) to IFS-ENS (dashed line); (3) Continuous Ranked Probability Score (CRPS), where lower values indicate better probabilistic skill; (4) Spread-Skill Ratio (SSR), where values near 1.0 indicate well-calibrated ensemble spread; (5) SDL-WXFormer rank histogram frequency showing distribution of verification values among ranked ensemble members; (6) IFS-ENS rank histogram frequency for comparison. The two RMSE rows share the same $y$-axis range within each column for direct comparison between deterministic and ensemble performance. Colors in rank histograms denote forecast lead times: 6~h (purple), 5~days (blue), 10~days (orange), 15~days (green).}
    \label{fig:pressure_levels}
\end{figure}

\begin{figure}[htbp]
    \centering
    \includegraphics[width=\columnwidth]{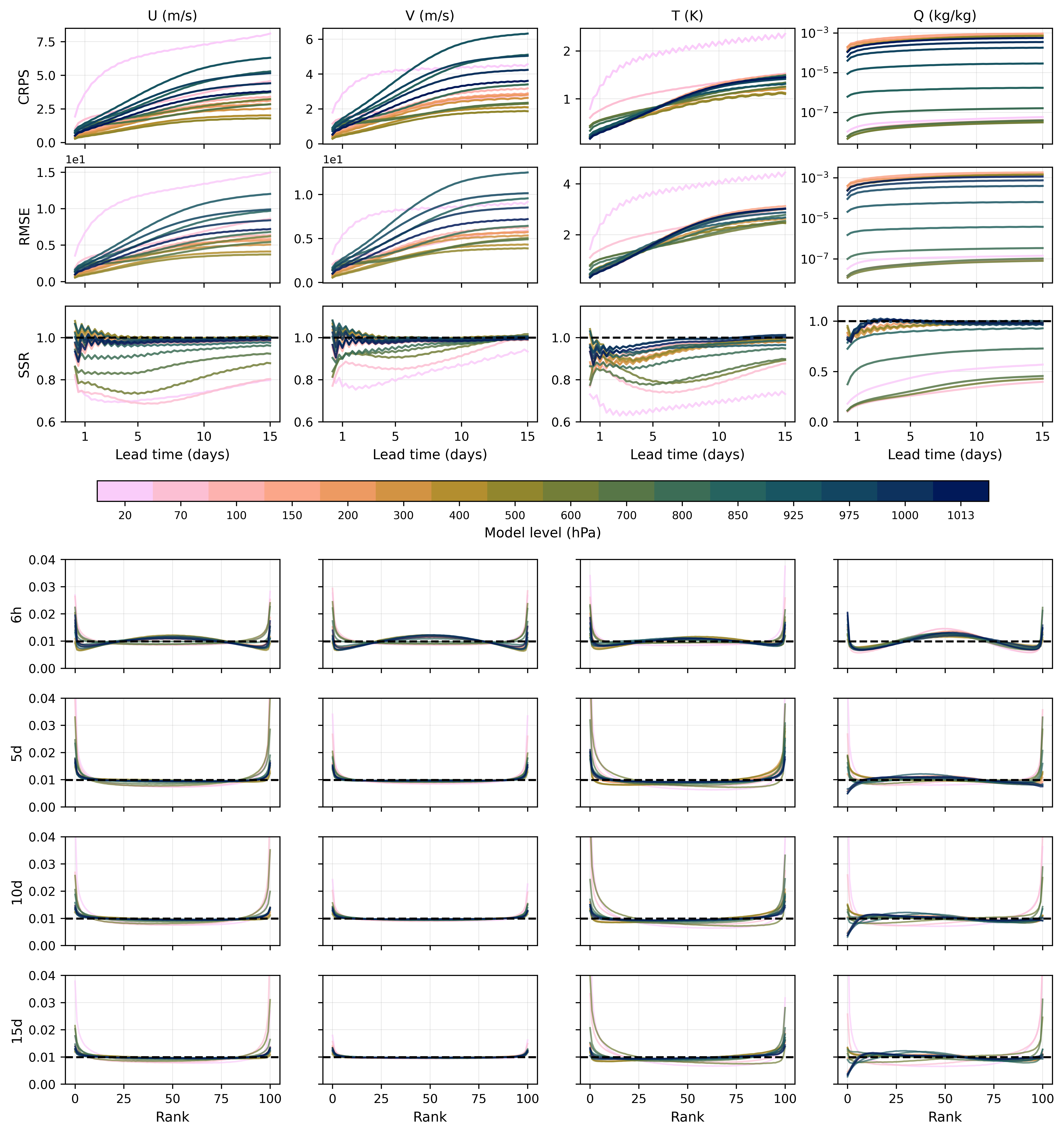}
    \caption{Vertical structure of ensemble forecast skill for core atmospheric variables across all 16 model levels. Columns show zonal wind (U), meridional wind (V), temperature (T), and specific humidity (Q). Rows display: (1-2) CRPS and RMSE as functions of forecast lead time (x-axis) and model level (colorbar, with lighter colors indicating upper atmosphere); (3) Spread-Skill Ratio (SSR) versus lead time, with dashed horizontal line at 1.0 indicating perfect calibration; (4-7) Rank histograms at 6h, 5d, 10d, and 15d lead times across ensemble member ranks (x-axis). Each rank histogram curve corresponds to a different model level (colored by vertical position). SDL-WXFormer (solid) and IFS-ENS (dashed) shown for SSR comparison. Model levels are hybrid $\sigma$--pressure coordinates; approximate pressure equivalents are provided for reference (e.g., level~90 $\approx$ 500~hPa, level~110 $\approx$ 850~hPa).
}
    \label{fig:vertical_profiles}
\end{figure}

We evaluate our SDL-enhanced WXFormer (SDL-WXFormer) ensemble against the ECMWF Integrated Forecasting System (IFS-ENS) on the 2022 test period using standard probabilistic metrics: Continuous Ranked Probability Score (CRPS), root mean square error (RMSE) of the ensemble mean, spread-skill ratio (SSR), and rank histograms. We also compare the underlying deterministic SDL-WXFormer against IFS HRES to establish baseline forecast skill. All forecasts are verified on a common 192×288 grid; details on operational system resolutions, regridding procedures, and pressure-level interpolation are provided in \ref{sec:skill_details}. Figure~\ref{fig:pressure_levels} shows results for key atmospheric variables at selected pressure levels, while Figure~\ref{fig:vertical_profiles} presents comprehensive vertical structure across all 16 model levels.

The SDL ensemble achieves competitive CRPS scores across all variables and lead times (Figures~\ref{fig:pressure_levels} and \ref{fig:vertical_profiles}, top rows). For upper-air winds (U700, V850), geopotential height (Z500), and temperature (T850), CRPS remains comparable to IFS-ENS through 15-day forecasts. Notably, specific humidity (Q850) demonstrates particularly strong performance with lower CRPS than IFS-ENS throughout the forecast window, indicating effective representation of boundary layer moisture processes. The vertical profiles reveal stratification in skill—lower-to-mid tropospheric variables (model levels 60--137) exhibit substantially lower CRPS and RMSE than upper tropospheric and stratospheric levels (model levels 10--40), reflecting the model's training on 16 selected model levels that emphasize weather-relevant atmospheric regions while accepting coarser vertical resolution at higher altitudes.

Deterministic RMSE comparisons (Figure \ref{fig:pressure_levels}, first row) show that the base deterministic WXFormer model delivers performance comparable to operational IFS HRES forecasts across all evaluated variables. For dynamical fields (Z500, U700, V850) and the thermodynamic field (T850), the model’s RMSE curves closely track—or in some cases outperform—IFS HRES throughout the 10-day IFS forecast window. Beyond day 10, the model maintains stable skill through day 15, demonstrating that the ERA5-trained dynamics extend reliably into a longer forecast horizon than that provided by IFS in this comparison. Specific humidity (Q850) shows particularly impressive deterministic skill, with consistently lower RMSE than IFS HRES across all lead times. This strong performance on moisture fields reflects effective learning of boundary layer processes and moisture transport despite the well-known challenges of humidity prediction in data-driven models \cite{bi2023pangu, Lam2023graphcast}.

Ensemble mean RMSE (Figure~\ref{fig:pressure_levels} and \ref{fig:vertical_profiles}, second rows) follows similar patterns to both deterministic RMSE and CRPS, with SDL forecasts maintaining accuracy comparable to or better than IFS-ENS through medium-range timescales. The shared $y$-axis range between deterministic and ensemble RMSE rows reveals that ensemble averaging provides consistent error reduction across all variables, with ensemble mean performance closely matching single-member deterministic skill. The RMSE growth rates are consistent with ensemble mean forecast error accumulation in operational systems, suggesting that the pre-trained deterministic WXFormer base provides stable skill while SDL perturbations add calibrated uncertainty without degrading deterministic performance.

\subsection{Ensemble Calibration}

Figure~\ref{fig:pressure_levels} shows calibration diagnostics for key pressure levels. The spread-skill ratio (SSR) quantifies ensemble calibration by comparing ensemble spread (standard deviation) to ensemble mean RMSE; well-calibrated ensembles exhibit SSR near 1.0. For all evaluated variables, SSR values start below unity at short lead times but converge toward optimal calibration (SSR $\approx$ 1.0) by day 5-10 and maintain well-calibrated spread through day 15. Specific humidity (Q850) demonstrates particularly strong calibration throughout the forecast window.

Rank histograms provide complementary calibration diagnostics by comparing the frequency with which verification values fall into ranked ensemble member bins. Ideally, rank histograms should be flat (uniform), indicating ensemble members and observations are statistically indistinguishable samples from the same distribution. All variables exhibit U-shaped patterns at short lead times (6h) indicating initial underdispersion, but these histograms progressively flatten toward uniform distributions by day 10-15 as ensemble spread grows with forecast range. IFS-ENS exhibits similar temporal evolution in rank histogram structure.

Figure~\ref{fig:vertical_profiles} reveals that calibration quality degrades with altitude. Lower-tropospheric levels (model levels 100-137) achieve SSR values near unity and produce substantially flatter rank histograms across all lead times compared to upper-tropospheric levels (model levels 10-60), which show persistent underdispersion (SSR $< 1.0$) and more pronounced U-shapes throughout the forecast window. This vertical gradient reflects the model's 16-level discretization, which provides finest resolution in the weather-relevant lower troposphere while accepting coarser sampling aloft.

\subsection{Ensemble Forecast Evolution Across Lead Times}

\begin{figure}[htbp]
\centering
\includegraphics[width=\textwidth]{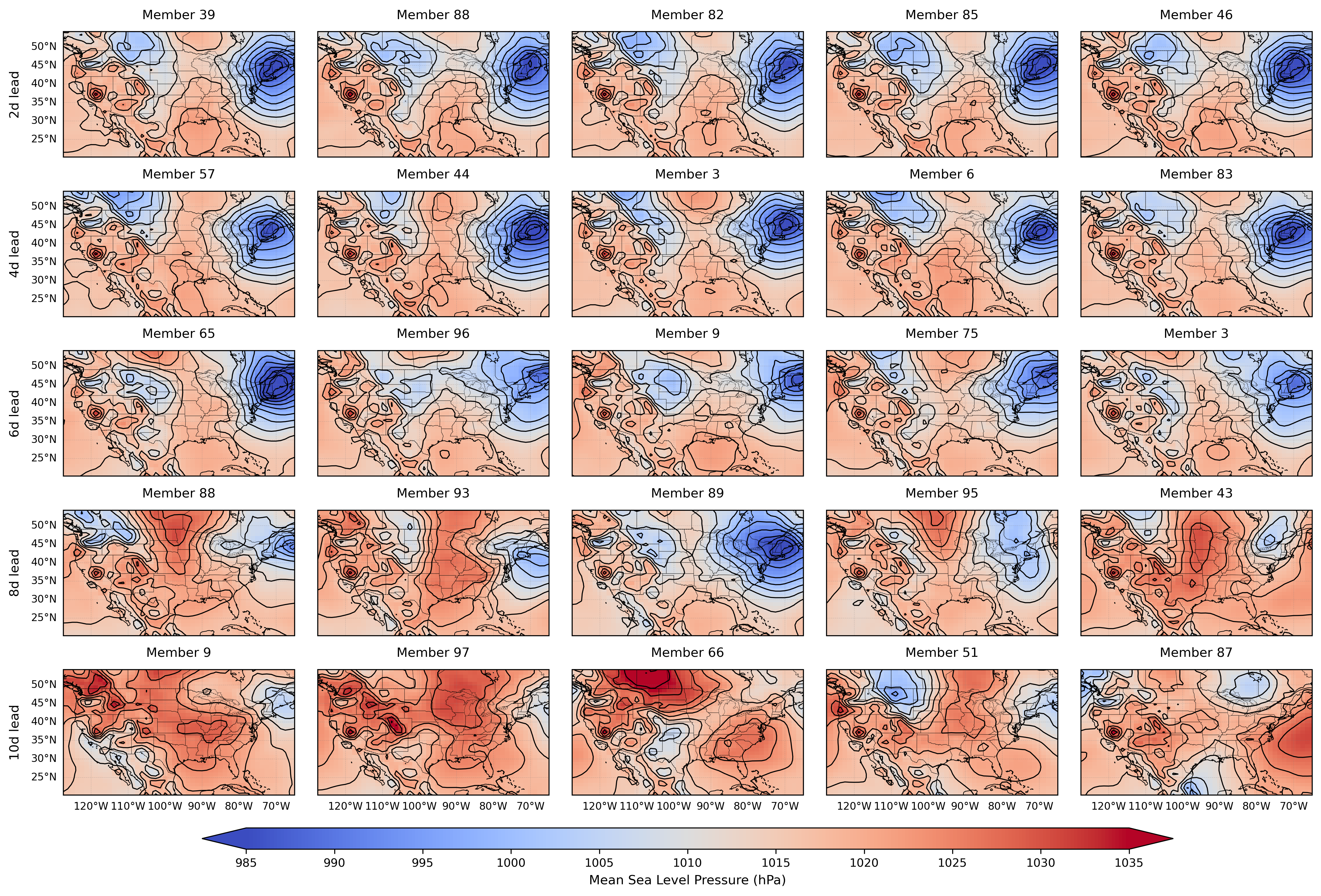}
\caption{Ensemble forecast evolution for Winter Storm Izzy valid January 17, 2022 at 18:00 UTC. Each row represents a different forecast initialization time (2-10 days before the event). For each lead time, five ensemble members are randomly shown. Contours show MSLP at 10 hPa intervals.}
\label{fig:izzy_ensemble}
\end{figure}

At 10-day lead time, the ensemble exhibits substantial uncertainty in both storm track and intensity (Figure~\ref{fig:izzy_ensemble}). The five randomly selected members span a wide range of solutions: some show weak offshore systems with minimal coastal impacts, others depict intense coastal lows, and several fail to capture organized cyclogenesis. Storm position varies by several hundred kilometers, with central pressure estimates spanning $>$ 30 hPa across the ensemble. This spread reflects fundamental uncertainty in synoptic patterns at extended range. At 6-day lead time, ensemble convergence is evident. All members depict well-defined coastal cyclogenesis, though disagreement persists regarding exact track and intensity. Spatial spread narrows considerably, with most members placing the system over the western Atlantic offshore of the Mid-Atlantic coast.

At 2-day lead time, the ensemble has strong agreement on storm structure and position. All sampled members show consistent development of an intense low pressure system (central pressures $<$ 980 hPa) positioned offshore of the Northeast coast. Remaining uncertainty concerns fine-scale intensity details and exact offshore positioning rather than fundamental occurrence or track. This evolution from high uncertainty to consensus illustrates typical improvement in forecast skill as lead time decreases for winter coastal cyclones, with ensemble spread collapsing as initial condition uncertainty diminishes and atmospheric predictability increases at shorter ranges.

\subsection{Reproducibility}

Operational forecast centers require the ability to exactly reproduce disseminated
ensemble members for verification, case studies, and audit trails. While stochastic perturbation methods can reproduce prior forecasts by storing random number generator (RNG) seeds, this approach only enables rigid replay. In contrast, SDL's latent tensor storage enables structural reproducibility, where the stored representation can be mathematically manipulated (e.g., rescaled or interpolated) post-inference. This is a capability unavailable with simple RNG seeds. SDL generates each ensemble member from three independent latent tensors $\mathbf{Z}_1, \mathbf{Z}_2, \mathbf{Z}_3 \sim \mathcal{N}(0, \mathbf{I})$ sampled at each decoder injection point. Rather than storing the raw latent vectors, we archive the computed noise perturbations $\alpha \cdot \mathbf{R}_{\text{Gauss}} \odot \mathbf{S}(\mathbf{Z}_\ell) \odot \mathbf{M}$ applied at each SDL layer during the forward pass. These stored perturbations are compact ($\sim$5~MB per member per forecast time), substantially reducing archival requirements relative to complete atmospheric state fields ($\sim$35~MB per member per time). Crucially, this representation supports post-inference manipulation through mathematical operations such as scaling or interpolation. This is a capability that distinguishes SDL from methods that discard noise after inference. 

\begin{figure}[ht]
\centering
\includegraphics[width=\textwidth]{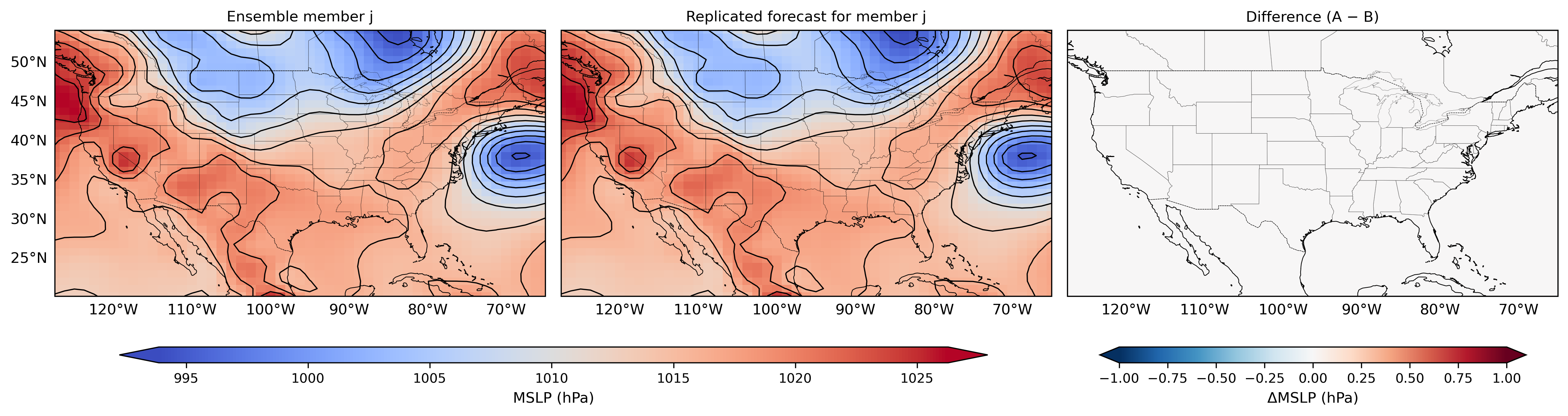}
\caption{Original ensemble member (left), regenerated forecast using stored latent tensor Z (center), and difference field (right) demonstrate exact reproduction and post-inference manipulation capability. The difference field shows maximum
absolute error below 0.01 hPa, confirming that storing the latent tensor enables perfect regeneration of any ensemble member.}
\label{fig:reproducibility}
\end{figure}

\subsection{Multi-Scale Uncertainty Control}
\label{sec:multi_scale_control}


\begin{figure}[htbp]
\centering
\includegraphics[width=0.95\textwidth]{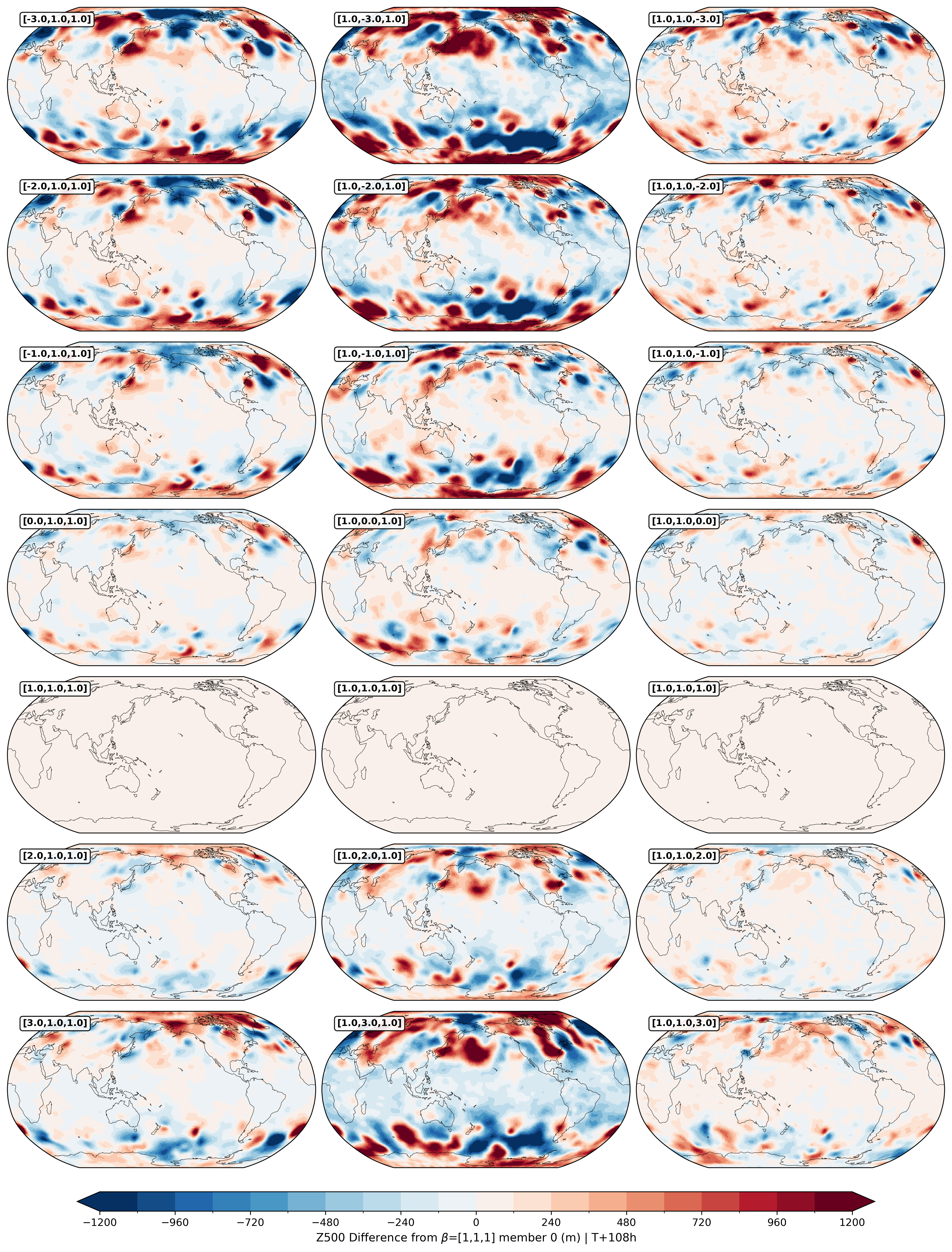}
\caption{Z500 anomalies from baseline configuration ($\beta=[1,1,1]$, member 0) demonstrating hierarchical control through single-parameter latent scaling. Columns represent systematic variation of $\beta_1$ (left), $\beta_2$ (center), and $\beta_3$ (right) across the range $[-3, 3]$ while holding remaining parameters at 1.0. Valid January 17, 2022 at 18:00 UTC (108h lead time).}
\label{fig:multiscale_control}
\end{figure}

\begin{figure}[htbp]
\centering
\includegraphics[width=\textwidth]{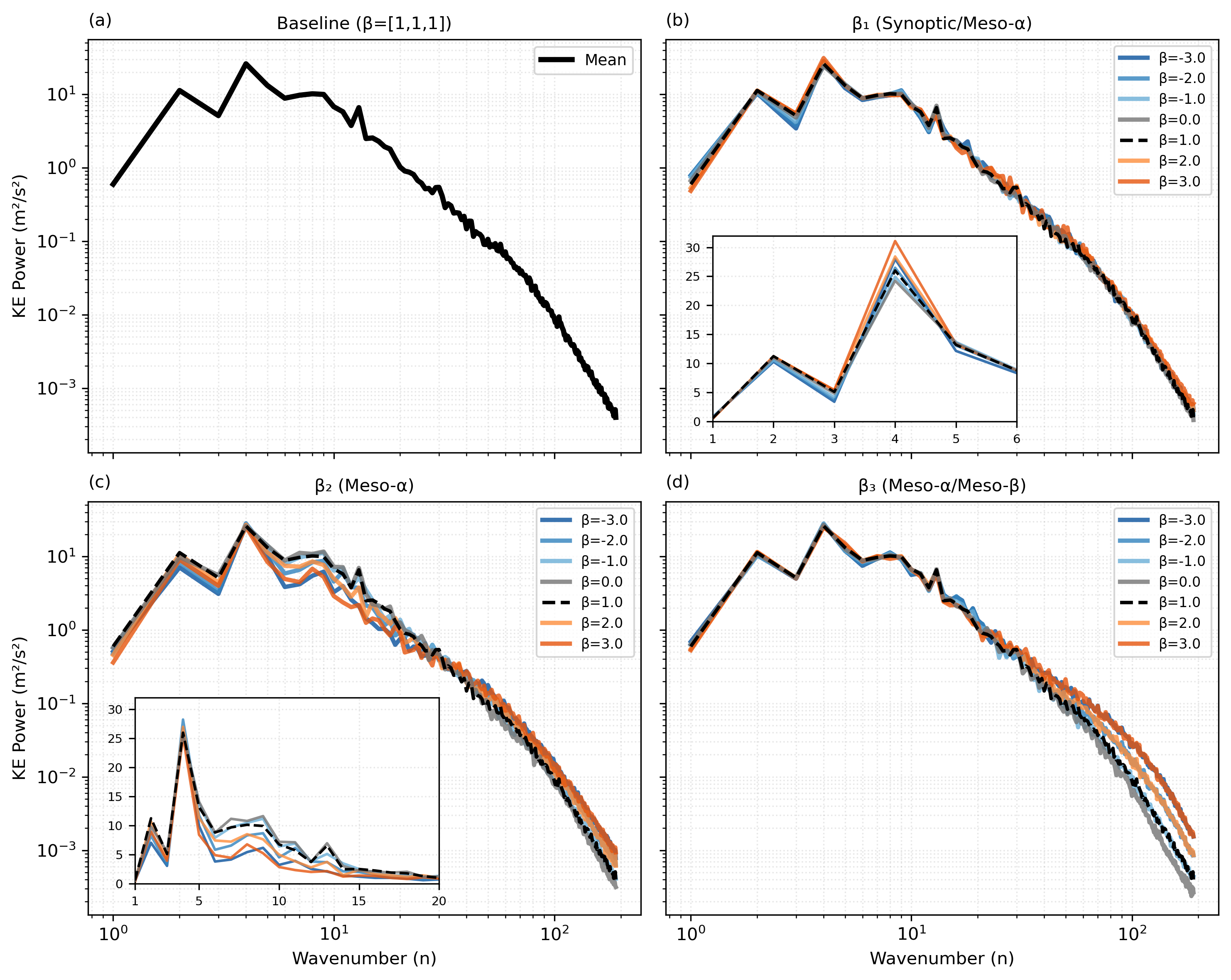}
\caption{Kinetic energy power spectra for systematic $\beta$ parameter variation, valid January 17, 2022 at 18:00 UTC (108h lead time) using 20-member ensembles with global spatial averaging across all vertical levels. Panel (a) displays baseline spectrum ($\beta=[1,1,1]$). Panels (b-d) present spectra for $\beta_1$, $\beta_2$, and $\beta_3$ variations across the range $[-3, 3]$ (colored lines) with remaining parameters fixed at 1.0. Black dashed line denotes $\beta$ = 1 reference configuration. Inset panels in (b-c) provide magnified views of wavenumber ranges 2-6 and 1-20 respectively.}
\label{fig:ke_spectra}
\end{figure}

We investigate whether SDL learns scale-dependent uncertainty decomposition aligned with the architectural hierarchy described in Table~\ref{tab:decoder_scales} through systematic manipulation of individual injection point contributions. During training, latent tensors are sampled from the standard normal distribution $\mathbf{Z} \sim \mathcal{N}(0, \mathbf{I})$ without additional scaling, which we denote as the $\beta = 1$ configuration. This establishes the nominal ensemble spread learned through CRPS optimization. At inference, we can manipulate this learned uncertainty structure by applying layer-specific scaling $\mathbf{Z}_\beta = \{\beta_1\mathbf{Z}_1, \beta_2\mathbf{Z}_2, \beta_3\mathbf{Z}_3\}$, where $\beta_1, \beta_2, \beta_3$ are independent scaling factors influencing the coarsest (Level 1), intermediate (Level 2), and finest (Level 3) injection points respectively. This formulation enables targeted amplification or suppression of perturbations at specific meteorological scales. To isolate individual layer contributions, we systematically vary one parameter across the range $[-3.0, 3.0]$ while holding remaining parameters fixed at their nominal training value of 1.0.


We analyze a global forecast valid January 17, 2022 at 18:00 UTC (108-hour lead time), corresponding to the mature phase of Winter Storm Izzy. For each parameter configuration, we generate ensemble member 0 using identical baseline noise but scaled latent tensors, computing Z500 anomalies relative to the baseline $\beta=[1,1,1]$ configuration. Figure~\ref{fig:multiscale_control} presents the systematic single-parameter exploration across 21 configurations spanning seven values per parameter. The anomaly framework isolates the direct meteorological impact of latent scaling by removing the dominant synoptic pattern common to all configurations.

Varying $\beta_1$ (left column of Figure~\ref{fig:multiscale_control}) while holding $\beta_2 = \beta_3 = 1.0$ reveals the coarsest layer's control over large-scale organization at the synoptic/meso-$\alpha$ boundary (Table~\ref{tab:decoder_scales}, Level~1: 800~km effective resolution). Z500 anomalies exhibit large-amplitude, spatially coherent patterns with magnitudes exceeding $\pm$600~m concentrated in mid-latitudes. Kinetic energy spectra (Figure~\ref{fig:ke_spectra}b) confirm that $\beta_1$ variations primarily modulate energy at synoptic scales and the large-scale end of the meso-$\alpha$ range (wavenumbers $n=2$--10), with the inset panel revealing complex, wavenumber-dependent energy redistribution: at $n=1$, $\beta=-3$ maximally increases KE; conversely, at the spectral peak ($n \approx 4$), $\beta=3$ generates the largest KE amplification. This wavenumber-dependent response demonstrates that coarse-layer perturbations redistribute energy across synoptic and large meso-$\alpha$ scales rather than applying uniform scaling.

The center column explores $\beta_2$ variation with $\beta_1 = \beta_3 = 1.0$, isolating intermediate-layer contributions at the meso-$\alpha$ scale (Level~2: 400~km, Table~\ref{tab:decoder_scales}). Z500 anomalies show comparable mid-latitude amplitudes to $\beta_1$ but with different spatial organization reflecting the intermediate architectural scale. Spectral analysis (Figure~\ref{fig:ke_spectra}c) reveals that $\beta_2$ produces the broadest influence across the wavenumber spectrum: all $\beta_2$ values suppress KE at the largest scales ($n=1$--3) relative to baseline, while generating substantial modulation extending from synoptic scales through the meso-$\alpha$ range ($n=4$--20+). The inset panel demonstrates peak sensitivity at intermediate wavenumbers ($n=10$--20), consistent with this layer's meso-$\alpha$ effective resolution. This broad spectral footprint explains why $\beta_2$ variations produce the most pronounced global Z500 anomaly patterns despite operating at an intermediate architectural level.

Right column results demonstrate that $\beta_3$ variation (finest layer, Level~3: 200~km at the meso-$\alpha$/meso-$\beta$ transition, Table~\ref{tab:decoder_scales}) produces minimal Z500 anomalies, typically below $\pm$200~m with limited spatial coherence. Spectral analysis (Figure~\ref{fig:ke_spectra}d) shows that $\beta_3$ effects emerge primarily at high wavenumbers ($n>20$--30) corresponding to meso-$\beta$ processes, with negligible impact on synoptic and large meso-$\alpha$ scales. This scale separation is consistent with the architectural design: perturbations at $2\times$ downsampling operate at scales below those dominating upper-tropospheric height fields, which are controlled by synoptic-scale and large meso-$\alpha$ wave dynamics.

While each SDL layer exhibits sensitivity across multiple wavenumbers—reflecting the overlapping receptive fields inherent to convolutional architectures—the spectral and spatial analyses confirm hierarchical stratification aligned with architectural scale. The $\beta_1$ layer produces maximum kinetic energy modulation at synoptic scales ($n=2$--10), $\beta_2$ dominates at meso-$\alpha$ scales ($n=10$--20), and $\beta_3$ effects emerge primarily at meso-$\beta$ scales ($n>20$--30). This scale-dependent behavior parallels findings from physics-based ensemble systems, where forecast variance evolution exhibits strong scale dependence with rapid saturation at meso-$\beta$ scales and slower growth at synoptic and large meso-$\alpha$ scales \cite{rodwell2025}.

\subsection{Latent Interpolation Between Ensemble Members}

\begin{table}[h!]
\centering
\small
\begin{tabular}{lccccccc}
\toprule
\textbf{$T$} & $e{=}0.0$ & $0.1$ & $0.2$ & $\dots$ & $0.8$ & $0.9$ & $1.0$ \\
\midrule
0 & $\mathbf{Z}_i(0)$ & $0.9\mathbf{Z}_i{+}0.1\mathbf{Z}_j$ & $0.8\mathbf{Z}_i{+}0.2\mathbf{Z}_j$ & $\dots$ & $0.2\mathbf{Z}_i{+}0.8\mathbf{Z}_j$ & $0.1\mathbf{Z}_i{+}0.9\mathbf{Z}_j$ & $\mathbf{Z}_j(0)$ \\
6h & $\mathbf{Z}_{e=0.0}(6\text{h})$ & $\mathbf{Z}_{e=0.1}(6\text{h})$ & $\mathbf{Z}_{e=0.2}(6\text{h})$ & $\dots$ & $\mathbf{Z}_{e=0.8}(6\text{h})$ & $\mathbf{Z}_{e=0.9}(6\text{h})$ & $\mathbf{Z}_{e=1.0}(6\text{h})$ \\
12h & $\mathbf{Z}_{e=0.0}(12\text{h})$ & $\mathbf{Z}_{e=0.1}(12\text{h})$ & $\mathbf{Z}_{e=0.2}(12\text{h})$ & $\dots$ & $\mathbf{Z}_{e=0.8}(12\text{h})$ & $\mathbf{Z}_{e=0.9}(12\text{h})$ & $\mathbf{Z}_{e=1.0}(12\text{h})$ \\
$\vdots$ & $\vdots$ & $\vdots$ & $\vdots$ & $\ddots$ & $\vdots$ & $\vdots$ & $\vdots$ \\
$T_f$ & $\mathbf{Z}_{e=0.0}(T_f)$ & $\mathbf{Z}_{e=0.1}(T_f)$ & $\mathbf{Z}_{e=0.2}(T_f)$ & $\dots$ & $\mathbf{Z}_{e=0.8}(T_f)$ & $\mathbf{Z}_{e=0.9}(T_f)$ & $\mathbf{Z}_{e=1.0}(T_f)$ \\
\bottomrule
\end{tabular}
\caption{Latent interpolation between initial states $\mathbf{Z}_i(0)$ and $\mathbf{Z}_j(0)$ with autoregressive evolution to future forecast times $T$. Each column shows an interpolated latent trajectory.}
\label{tab:latent_interp}
\end{table}

The deterministic latent space enables controlled exploration between ensemble members. Given two stored latent tensors $\mathbf{Z}_1$ and $\mathbf{Z}_2$ from different members, we generate intermediate forecasts:
\begin{equation}
\label{eq:latent_interp}
\mathbf{Z}_e = (1-e)\mathbf{Z}_1 + e\mathbf{Z}_2, \quad e \in [0, 1]
\end{equation}
Table~\ref{tab:latent_interp} shows that interpolation applies independently at each forecast step. For every parameter $e \in [0,1]$, the latent state $\mathbf{Z}_e(T)$ is a weighted combination of $\mathbf{Z}_i(T)$ and $\mathbf{Z}_j(T)$.

\begin{figure}[t]
\centering
\includegraphics[width=\textwidth]{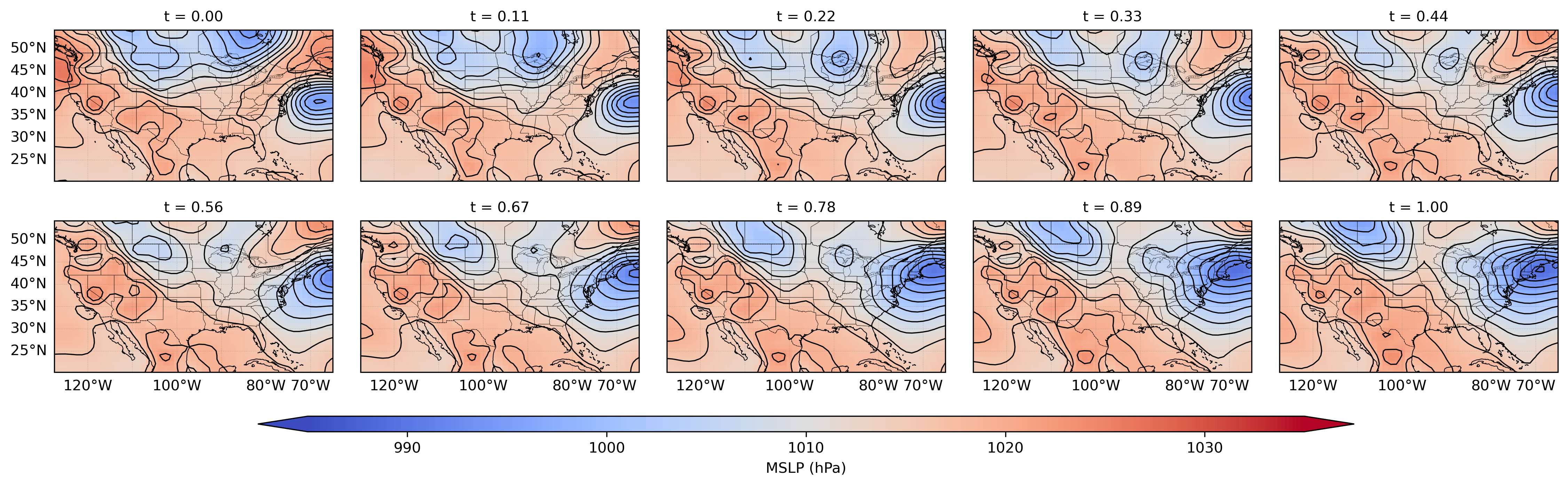}
\caption{Ten forecasts along the interpolation path between two ensemble members ($e = 0$ and $e = 1$) demonstrate smooth transitions in storm structure.}
\label{fig:latent_interp}
\end{figure}

Figure~\ref{fig:latent_interp} shows ten interpolated forecasts for Winter Storm Izzy at 6.5-day lead time. At $e{=}0.00$, the forecast shows a broad trough over the central United States with a weak coastal disturbance. At $e{=}1.00$, the forecast shows a mature extratropical cyclone positioned offshore. The intermediate forecasts transition smoothly. The western trough weakens and shifts east. The coastal system intensifies and migrates northeast. Pressure gradients tighten as the offshore low deepens. Storm structure, trough orientation, and circulation patterns adjust systematically. Linear paths through SDL's latent space produce dynamically consistent meteorological transformations.

\subsection{Uniform Ensemble Spread Calibration}

Multi-scale perturbation experiments (Section~\ref{sec:multi_scale_control}) demonstrate that individual SDL injection points exert independent control over uncertainty at different spatial scales, enabling targeted adjustments of synoptic versus mesoscale spread. While this hierarchical control provides scientific insight into learned uncertainty structures, operational recalibration workflows benefit from simpler parameterizations. Uniform scaling, where $\beta_1 = \beta_2 = \beta_3 = \beta$, reduces the adjustment space to a single interpretable parameter representing overall ensemble dispersion. Given stored latent tensor $\mathbf{Z}$, applying $\mathbf{Z}_{\beta} = \beta \mathbf{Z}$ uniformly across all SDL injection points enables post-inference spread adjustment without forecast regeneration.

\begin{figure}[b!]
\centering
\includegraphics[width=\textwidth]{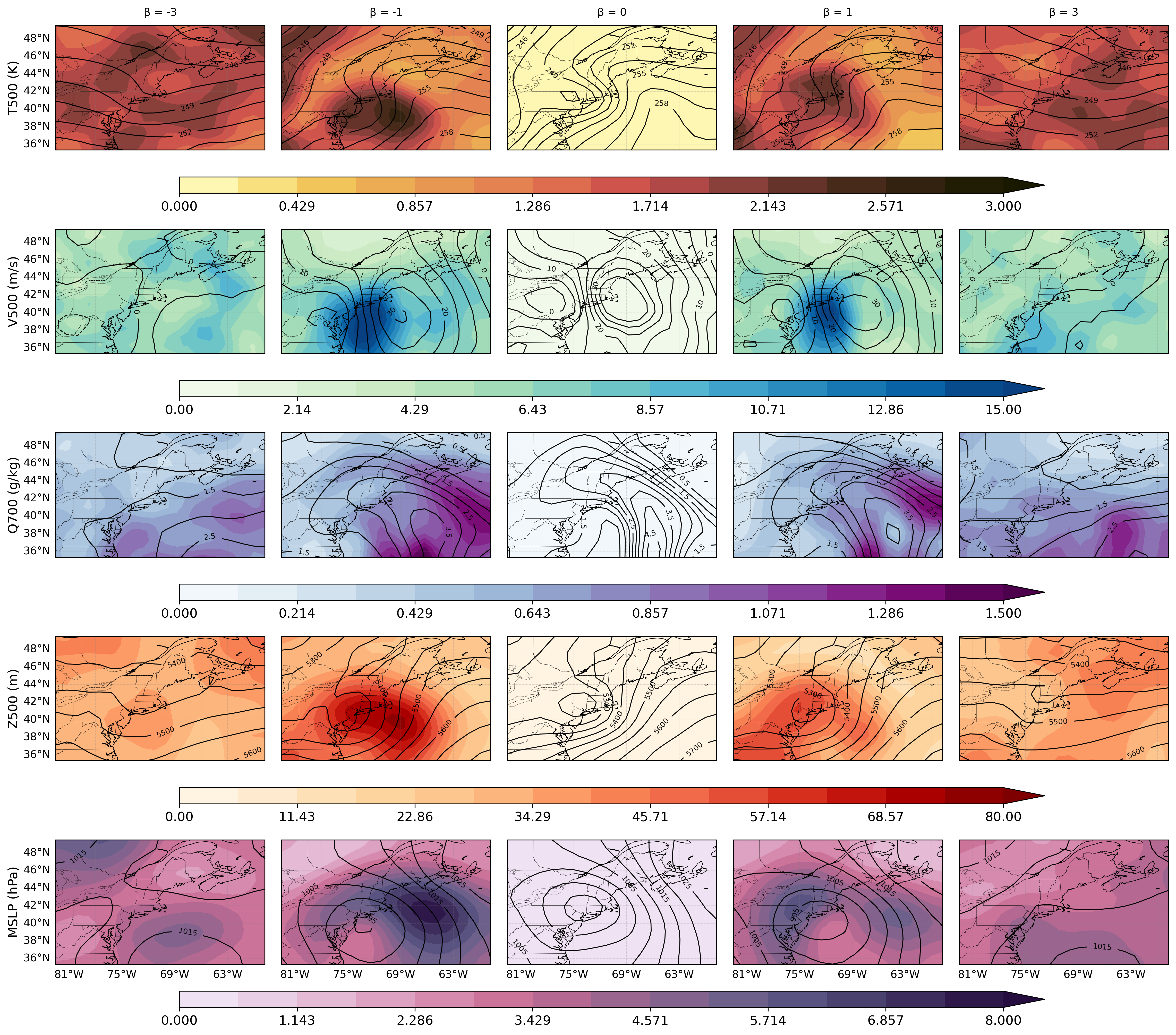}
\caption{Ensemble standard deviation across five atmospheric variables for Winter Storm Izzy at 108-hour lead time as a function of latent scaling factor $\beta$. Rows show T500, V500, Q700, Z500, and MSLP. Black contours depict ensemble mean fields. Each panel represents 20-member ensemble spread with uniform scaling $\beta_1 = \beta_2 = \beta_3 = \beta$ applied across all decoder injection points.}
\label{fig:spread_control}
\end{figure}

Figure~\ref{fig:spread_control} quantifies the relationship between uniform latent scaling and ensemble spread for Winter Storm Izzy at 108-hour lead time. Five atmospheric variables demonstrate systematic spread modulation across $\beta \in [-3, 3]$, with uncertainty maxima concentrating in baroclinic zones, jet streams, and cyclone centers. The symmetric behavior with respect to $\beta = 0$ is clear. At the training distribution reference ($\beta = 1$), spread patterns align with dynamically expected structures and medium-range forecast error climatology. Both positive and negative amplification ($|\beta| = 3$) extend spread uniformly while maintaining spatial organization and physical realism.

\begin{figure}[b!]
\centering
\includegraphics[width=\textwidth]{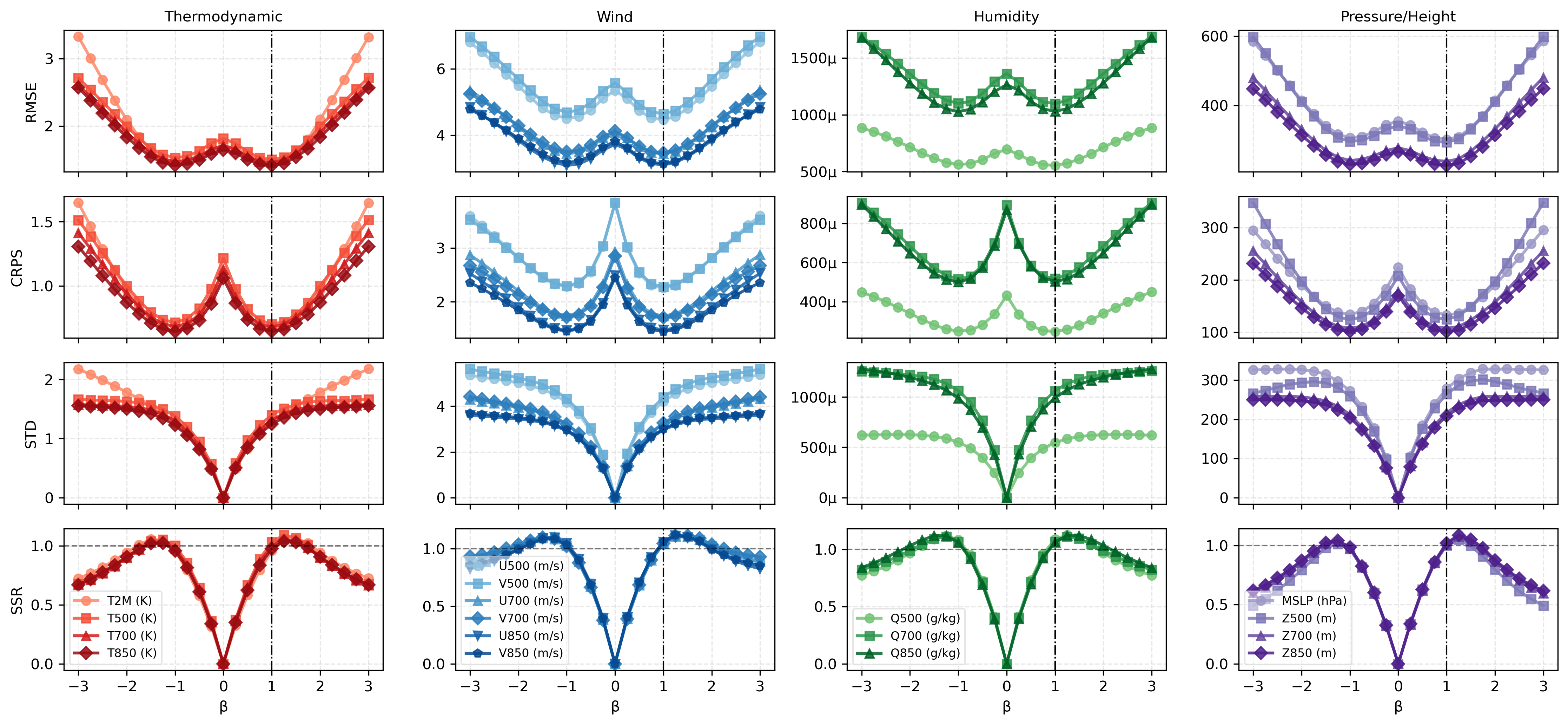}
\caption{Verification metrics versus uniform latent scaling factor $\beta \in [-3, 3]$ for the globe at 108-hour lead time. Rows show RMSE, CRPS, ensemble standard deviation (STD), and spread-skill ratio (SSR) across four atmospheric variable categories. Vertical dashed line at $\beta$ = 1.0 marks nominal training distribution. Note symmetric structure: both $\beta$ = $-1$ and $\beta$ = $+1$ produce well-calibrated ensembles with SSR $\approx 1.0$.}
\label{fig:latent_scaling_metrics}
\end{figure}

Ensemble mean fields (black contours) vary systematically with $\beta$ scaling while preserving overall synoptic structure and demonstrating symmetric behavior with respect to $\beta$ = 0. Contour patterns at $\beta$ = -3 and $\beta$ = +3 exhibit similarity, as do configurations at $\beta$ = -1 and $\beta$ = +1, confirming that uniform latent scaling modulates both spread and ensemble central tendency in a geometrically symmetric manner. The preservation of dynamically consistent spatial patterns across the full range $|\beta| \leq 3$, where spread amplitudes deviate substantially from training distribution values, suggests that SDL's learned perturbation structure captures fundamental uncertainty relationships rather than merely fitting statistical properties of the training distribution. Quantitative verification of this symmetric scaling behavior is presented in Figure~\ref{fig:latent_scaling_metrics}.

Figure~\ref{fig:latent_scaling_metrics} reveals symmetric behavior with respect to $\beta = 0$, demonstrating a fundamental property of SDL's learned latent space. Ensemble standard deviation (third row) scales monotonically with $|\beta|$, approaching zero at $\beta = 0$ (deterministic forecast) and increasing approximately linearly for $|\beta| < 1.5$ across all variable categories. Critically, $\beta = -1$ produces ensemble spread nearly identical to $\beta = +1$, indicating that negative scaling inverts perturbation direction while maintaining equivalent uncertainty magnitudes. This symmetry demonstrates that the learned latent space supports bidirectional exploration of forecast uncertainty around the deterministic baseline. 

Spread-skill ratio (bottom row) achieves values near unity at both $\beta \approx -1$ and $\beta \approx +1$, revealing that the symmetric latent structure produces equivalently calibrated ensembles despite training exclusively with positive latent samples ($\beta = 1$). Well-calibrated ensembles (SSR within [0.8, 1.2]) persist across $|\beta| \in [0.75, 3.0]$ for all variable categories. Within the range explored, SSR degrades only as $\beta$ approaches zero, where ensemble spread collapses toward the deterministic limit while forecast error remains non-zero.

Ensemble mean RMSE (top row) exhibits U-shaped profiles symmetric around $\beta = 0$ with minima near $|\beta| \approx 1$ for all variable categories, confirming that forecast quality depends on perturbation magnitude rather than direction. For $|\beta| < 1.25$, RMSE remains within 10\% of minimum values across all variables. CRPS (second row) reveals optimal probabilistic performance near $|\beta| \approx 1$ with symmetric degradation as $|\beta|$ increases beyond 1.5 or decreases below 0.5. The smooth, predictable relationships between $|\beta|$ and ensemble spread enable operational post-hoc recalibration where archived latent tensors support retrospective spread adjustment without forecast regeneration, a capability currently unavailable in existing ML ensemble methods where stochastic perturbations are discarded after inference. Operational applications can select $|\beta| \approx 1.0$ for nominal uncertainty quantification, $|\beta| < 1.0$ for conservative spread estimates, or $|\beta| > 1.0$ for expanded phase space sampling, with equivalent calibration for both positive and negative perturbations.

\section{Discussion}

\subsection{Learned Uncertainty Structure}

SDL learns perturbation structure implicitly from ERA5 reanalysis through CRPS optimization, with no explicit physics encoded in the noise injection mechanism. ERA5 is generated by the IFS data assimilation system and inherits its uncertainty structure, including both physical variability and systematic model errors. SDL therefore learns to emulate IFS ensemble spread characteristics, spatial correlations, and trajectory divergence patterns present in the training data.

Several lines of evidence suggest SDL learns structured representations rather than merely memorizing statistical distributions. Linear latent interpolation produces physically realistic intermediate states: pressure systems intensify or weaken gradually, trough axes rotate smoothly, and frontal boundaries shift systematically. This suggests the latent space possesses geometric structure where directions in $Z$-space correspond to meaningful meteorological variations. The systematic relationship between latent scaling factor $\beta$ and ensemble spread provides further evidence of learned structure. During training, latent tensors are sampled from $Z \sim \mathcal{N}(0, I)$ with the CRPS loss enforcing ensemble spread calibration against ERA5 variability. When $\beta \neq 1$ is applied at inference, forecasts remain physically reasonable with predictable spread scaling despite extrapolating beyond the training distribution.

The learned latent space exhibits symmetric behavior with respect to $\beta = 0$, where opposite-sign perturbations produce equivalent ensemble dispersion. This is a property that emerged naturally from CRPS optimization without explicit architectural constraints. Well-calibrated ensembles (SSR within [0.8, 1.2]) persist across $|\beta| \in [0.75, 3.0]$ for all variable categories, representing threefold variations in ensemble dispersion relative to the training distribution. Calibration degrades only as $\beta$ approaches zero, where ensemble spread collapses toward the deterministic limit while forecast error remains non-zero. This symmetry enables operational flexibility: forecasters can explore alternative ensemble configurations through sign inversion, effectively doubling the available calibration states without additional computation.

SDL's uncertainty quantification was calibrated against ERA5 variability from 1979–2018, and these learned uncertainty bounds may not generalize to other time periods or future climate states. For unprecedented weather regimes or atmospheric conditions substantially outside the training distribution, SDL's ensemble may underrepresent tail risks. Evaluating ensemble spread appropriateness under such out-of-distribution conditions remains an open question requiring targeted validation. This contrasts with physics-based stochastic parameterization schemes that can potentially extrapolate to novel regimes through explicit representation of physical processes.

Additionally, the 16-level vertical discretization optimized for lower-tropospheric resolution limits representation of stratospheric processes relevant to medium-range prediction. Sudden stratospheric warmings, stratosphere-troposphere coupling during blocking events, and quasi-biennial oscillation teleconnections may be inadequately captured in upper levels where vertical resolution is coarsest. Applications requiring explicit stratospheric dynamics would benefit from expanded vertical discretization or hybrid approaches combining fine tropospheric resolution with targeted stratospheric sampling.

\subsection{Relationship to Physical Parameterization}

Stochastic parameterization in numerical weather prediction represents model uncertainty arising from unresolved scales and uncertain physics \cite{berner2017stochastic, palmer2009stochastic, palmer2019stochastic}. Machine learning approaches for stochastic parameterization include generative adversarial networks applied to simplified atmospheric models \cite{gagne2020machine} and comprehensive reviews of data-driven stochastic methods \cite{christensen2024machine}. These methods are designed based on theoretical understanding of atmospheric processes and can be interpreted through physical mechanisms. 

Despite this fundamental difference, SDL ensembles exhibit similar statistical properties to physics-based systems: calibrated spread-skill relationships, close-to uniform rank histograms, and realistic trajectory divergence. This suggests that the functional form of uncertainty, including how perturbations are structured across scales, variables, and forecast lead times, can be learned from data through appropriate training objectives. SDL never explicitly encodes knowledge about baroclinic instability or moist convection, yet it produces ensembles whose statistical behavior resembles systems that explicitly parameterize these processes through CRPS optimization on IFS-derived reanalysis data.

The learned nature of SDL has both advantages and limitations compared to theory-driven approaches. SDL can capture systematic model errors and complex uncertainty patterns that emerge from the full modeling system without requiring explicit specification. However, it lacks interpretability in terms of physical processes and cannot generalize beyond training distribution bounds. These approaches are complementary rather than competing. The learned perturbations could augment physics-based stochastic schemes by capturing emergent uncertainty patterns while retaining interpretable physical perturbations for fundamental atmospheric processes.

While SDL shares computational efficiency with other CRPS-trained ensemble methods, its hierarchical latent parameterization distinguishes it through explicit control mechanisms. The learned latent space enables targeted scale-dependent perturbation and post-inference spread adjustment—capabilities that provide operational flexibility beyond what is achievable with fixed stochastic injection schemes.


\subsection{Hierarchical Uncertainty Decomposition}

The multi-scale injection strategy produces scale-dependent uncertainty control that remains coherent even when individual layer perturbations are amplified beyond training distribution magnitudes. This robustness suggests the learned decomposition captures fundamental atmospheric uncertainty structure rather than distribution-specific artifacts. Critically, this hierarchical organization emerges from CRPS optimization without explicit architectural constraints enforcing scale separation, analogous to learned hierarchical representations in generative modeling \cite{karras2019style, karras2020analyzing} where stochastic injection combined with appropriate loss functions induces natural scale-dependent representations.

\section{Conclusions}

We introduce Stochastic Decomposition Layers, a computationally efficient method for converting deterministic machine learning weather models into calibrated probabilistic ensemble forecasting systems. SDL employs hierarchical noise decomposition to inject learned stochastic perturbations at multiple scales through latent-driven modulation, per-pixel noise, and learned channel modulation. When applied to WXFormer via transfer learning, SDL achieves three primary contributions.

First, training efficiency through transfer learning: fine-tuning from pre-trained deterministic weights requires approximately 1.61\% of the computational cost needed to train the baseline model from scratch. Although all model weights undergo joint optimization during fine-tuning, the pre-trained initialization dramatically accelerates convergence compared to random initialization. Inference exhibits linear scaling with ensemble size, matching CRPS-trained alternatives while avoiding the iterative refinement overhead of diffusion methods.

Second, probabilistic forecast quality: evaluation on 2022 ERA5 reanalysis demonstrates reliable ensemble calibration through medium-range forecasts, with ensemble spread properly representing forecast uncertainty across dynamical and thermodynamic variables. The model maintains superior skill in the lower troposphere where vertical resolution is finest, with greater challenges in the upper atmosphere where the 16-level discretization provides coarser vertical sampling.

Third, controllable uncertainty quantification: the explicit latent parameterization enables perfect reproducibility through compact latent storage, post-inference spread adjustment via latent rescaling, and systematic scenario exploration through latent interpolation. The learned latent space exhibits symmetric properties where opposite-sign perturbation scaling produces equivalent ensemble dispersion.

SDL's learned uncertainty structure is bounded by the ERA5 training distribution (1979-2022), and may underrepresent tail risks for unprecedented weather regimes or climate states substantially different from the training period. Future work should explore state-dependent noise injection, training on multiple reanalysis products to capture inter-model spread, and hybrid approaches combining learned perturbations with physics-based stochastic parameterization.

SDL demonstrates that hierarchical stochastic decomposition provides effective controllable generation for atmospheric prediction, offering a parameter-efficient alternative to diffusion-based methods while enabling post-inference control over ensemble spread and reproducibility through compact latent storage.

\appendix

\section{Baseline Model Training Methodology}
\label{sec:train_details}

\subsection{Deterministic Baseline Training Schedule}

The deterministic WXFormer baseline model underwent progressive multi-step training following established curriculum learning protocols for autoregressive weather prediction. Training employed ERA5 reanalysis data spanning 1979-2017 with the temporal split and preprocessing procedures described in Section 3.

The training curriculum comprised five sequential phases with progressively increasing rollout lengths to enable stable learning of long-horizon weather dynamics. Phase 1 employed single-step mean squared error (MSE) loss for 70 epochs using 1,781 batch updates per epoch with batch size 32. Subsequent phases utilized multi-step MSE loss with gradient computation at each autoregressive timestep to ensure proper temporal credit assignment throughout the rollout sequence.

Phase 2 implemented two-step autoregressive training for 20 epochs using 1,780 batch updates per epoch with batch size 32. Phase 3 extended to four-step rollouts for 20 epochs using 1,000 batch updates per epoch with batch size 32. Phase 4 employed eight-step rollouts for 20 epochs using 500 batch updates per epoch with batch size 32. Phase 5 concluded with sixteen-step rollouts for 20 epochs using 100 batch updates per epoch with batch size 32.

The systematic reduction in batch updates per epoch across phases accommodated memory constraints imposed by longer autoregressive sequences while maintaining consistent training sample coverage. Gradient backpropagation occurred at each timestep throughout all multi-step phases, ensuring comprehensive temporal gradient flow and stable convergence properties across extended forecast horizons.

\subsection{Ensemble Fine-Tuning Configuration}

SDL ensemble fine-tuning initialized from the fully trained deterministic baseline and optimized the introduced SDL parameters through joint fine-tuning of all model weights. Training employed single-step almost fair Continuous Ranked Probability Score (afCRPS) loss (Equation 2) for 20 epochs using 1,000 batch updates per epoch. Ensemble generation during training utilized batch dimension replication with effective batch size 10, corresponding to 10 ensemble members per input sample. This approach leveraged the pre-trained deterministic forecast skill as a strong initialization while requiring substantially reduced computational resources compared to training ensemble systems from random initialization.

\subsection{Computational Cost Assessment}

Computational cost comparison utilized forward and backward pass enumeration across all training phases. The complete deterministic baseline curriculum required 12,410,560 total forward passes distributed across: single-step training (3,988,160 passes), two-step training (2,278,400 passes), four-step training (2,560,000 passes), eight-step training (2,560,000 passes), and sixteen-step training (1,024,000 passes). Each forward pass required a corresponding backward pass for gradient computation, yielding equivalent totals for both operations.

SDL ensemble fine-tuning required 200,000 forward passes and an equivalent number of backward passes over the 20-epoch training period. Applying standard computational cost ratios for transformer architectures, where backward passes require approximately twice the computational expense of forward passes, the relative training cost of SDL fine-tuning compared to baseline training equals 1.61\% of total baseline computational requirements. This efficiency gain demonstrates the practical advantage of transfer learning approaches for ensemble weather prediction systems.

\section{Forecast Verification Methodology}
\label{sec:skill_details}

\subsection{Model Training and Native Resolutions}

SDL-WXFormer was trained on ERA5 reanalysis regridded to 192×288 horizontal resolution (approximately 0.94° latitude × 1.25° longitude) using 16 selected model levels from ERA5's native 137-level hybrid sigma-pressure vertical coordinate system. The model produces forecasts on these model levels, which are then interpolated to standard pressure levels (500, 700, 850 hPa) during evaluation using the pressure-level interpolation procedure described in \cite{schreck2025}. ERA5 verification data undergo identical model-to-pressure level interpolation to ensure consistent comparison.

The ECMWF IFS ensemble system (IFS-ENS) is accessed through WeatherBench 2 archives, which provides IFS-ENS forecasts regridded to a 240×121 equiangular grid. We use this WeatherBench 2 distribution format rather than the native IFS-ENS resolution (O640 octahedral reduced Gaussian grid, ~9 km) due to data storage constraints. The ECMWF high-resolution deterministic system (HRES) archived data are provided at 1440×721 resolution (0.25° regular latitude-longitude). Both operational systems provide output on pressure levels directly.

\subsection{Verification Grid Configuration}

All forecast verification in this study uses a common 192×288 grid. For ensemble forecast evaluation (Figures 2–3), IFS-ENS forecasts were bilinearly interpolated from the WeatherBench 2 distribution grid (240×121 equiangular) to the 192×288 regular latitude-longitude verification grid. ERA5 verification fields were similarly interpolated to 192×288 from their native resolution. SDL-WXFormer forecasts are produced natively at 192×288 resolution and require no horizontal interpolation.

For deterministic forecast comparison (Figure~\ref{fig:pressure_levels}, first row), HRES forecasts were interpolated from 1440×721 to the 192×288 verification grid. This substantial spatial coarsening from 0.25° to approximately 1° resolution applies implicit smoothing that may reduce small-scale error magnitudes. ERA5 verification data were regridded to the same 192×288 grid.

Following horizontal interpolation to the common grid, SDL-WXFormer predictions and ERA5 verification both undergo vertical interpolation from model levels to pressure levels. IFS-ENS and HRES forecasts are already provided on pressure levels and require no vertical interpolation.

\subsection{Metric Computation and Implications}

Forecast skill metrics (RMSE, MAE, ACC) are computed on the 192×288 verification grid using cosine-latitude weights normalized such that $\sum_{j} w_j = N_{\text{lat}}$ to account for meridional convergence. Ensemble metrics (CRPS, spread-skill ratio, rank histograms) use the almost-fair CRPS formulation (Equation 2) with $\alpha = 0.95$. The 2022 test period comprises 365 forecast initializations at 00Z.

The common verification grid enables direct metric comparison but introduces resolution-dependent effects. Coarsening HRES from 0.25° to 1° applies substantial spatial smoothing that suppresses small-scale features and may artificially reduce deterministic RMSE. However, effective resolution of numerical weather prediction models is typically 4–5 grid spacings due to numerical diffusion and spectral truncation, suggesting HRES effective resolution of approximately 1°–1.25° despite 0.25° nominal spacing. The 192×288 grid therefore captures the predictive information relevant for medium-range evaluation while enabling consistent comparison. Ensemble spread calibration (spread-skill ratios near unity, uniform rank histograms) indicates appropriate probabilistic skill at the scales resolved by this verification grid.

\section*{Open Research Section}
The ERA5 reanalysis data for this study can be accessed through the NSF NCAR Research Data Archive \cite{hersbach2020era5}. Operational IFS-HRES deterministic forecasts and IFS-ENS ensemble forecasts are obtained from the WeatherBench 2 archive \cite{rasp2023weatherbench2}, which provides regridded ECMWF operational predictions on standardized verification grids at 240×121 (ensemble) and 1440×721 (deterministic) horizontal resolutions. The neural networks described here and the simulation code used to train and test the models are archived at \url{https://github.com/NCAR/miles-credit}.

\section*{Acknowledgments}
This material is based upon work supported by the NSF National Center for Atmospheric Research, which is a major facility sponsored by the U.S. National Science Foundation under Cooperative Agreement No. 1852977. This research has also been supported by NSF Grant No. RISE-2019758. We would like to acknowledge high-performance computing support from Derecho and Casper \cite{Cheyenne} provided by the Computational and Information Systems Laboratory, NCAR, and sponsored by the National Science Foundation. 

\subsection*{Conflict of Interest Statement}
The authors have no conflicts of interest to disclose.

\bibliographystyle{unsrtnat}
\bibliography{references}

\end{document}